\documentclass[letterpaper, 10pt, journal,twoside]{IEEEtran}
\usepackage{amsmath,amssymb,euscript,yfonts,psfrag,latexsym,dsfont,graphicx}
\usepackage{bbm,color,amstext,wasysym,balance,cite}
\graphicspath{{./},{./figures/}}

\usepackage{microtype}
\usepackage{subcaption}
\usepackage{booktabs} 

\usepackage{hyperref}
\usepackage[normalem]{ulem}

\usepackage{hyperref}       
\usepackage{url}            
\usepackage{booktabs}       
\usepackage{amsfonts}       
\usepackage{nicefrac}       
\usepackage{algorithm}
\usepackage{algorithmic}
\usepackage{caption}
\usepackage{xcolor}

\newtheorem{thm}{Theorem}
\newtheorem{rem}{Remark}

\newcommand{\bn}{{\mathbf n}}
\newcommand{\bo}{{\mathbf o}}
\newcommand{\by}{{\mathbf y}}

\newcommand{\citet}{\cite}

\newcommand{\bx}{{\bf x}}

\def\minwrt[#1]{\underset{#1}{\text{minimize}}}
\def\maxwrt[#1]{\underset{#1}{\text{maximize }}}

\IEEEoverridecommandlockouts

\begin{document}
\title{Incremental inference of collective graphical models}

\author{Rahul Singh, Isabel Haasler, Qinsheng Zhang, Johan Karlsson, and Yongxin Chen
\thanks{This work was supported by the Swedish Research Council (VR), grant 2014-5870, SJTU-KTH cooperation grant and the NSF under grant 1901599 and 1942523.
}
\thanks{R. Singh, Q. Zhang and Y. Chen are with the School of Aerospace Engineering,
Georgia Institute of Technology, Atlanta, GA, USA. {\tt\small \{qzhang419,rasingh,yongchen\}@gatech.edu}}
\thanks{I.~Haasler and J.~Karlsson are with the Division of Optimization and Systems Theory, Department of Mathematics, KTH Royal Institute of Technology, Stockholm, Sweden. {\tt\small haasler@kth.se, johan.karlsson@math.kth.se}}}

\maketitle
\thispagestyle{empty}
\begin{abstract}
We consider incremental inference problems from aggregate data for collective dynamics. In particular, we address the problem of estimating the aggregate marginals of a Markov chain from noisy aggregate observations in an incremental (online) fashion. We propose a sliding window Sinkhorn belief propagation (SW-SBP) algorithm that utilizes a sliding window filter of the most recent noisy aggregate observations along with encoded information from discarded observations. Our algorithm is built upon the recently proposed multi-marginal optimal transport based SBP algorithm that leverages standard belief propagation and Sinkhorn algorithm to solve inference problems from aggregate data. We demonstrate the performance of our algorithm on applications such as inferring population flow from aggregate observations. 
\end{abstract}

\section{Introduction}\label{sec:intro}
\IEEEPARstart{T}{he} problem of inference from aggregate data is widely studied in fields including machine learning, ecology, and social sciences~\cite{SheDie11, SunSheKum15}.
Similar problems also occur in the area of estimation and control, for instance in ensemble filtering \cite{Eve03, LorNae11,CheKar18,HasRinChe19}.
In these applications, one aims to infer information about a group of agents in the case where only aggregate observations in the form of counts or contingency tables are provided~\cite{SheDie11}. Information about individuals may not be available due to, e.g., economical or privacy reasons \cite{SheDie11}. 
For example, in bird migration analysis, individual trajectories are not readily accessible, but the number of birds in different areas can typically be counted from pictures. Another very present application is the spread of infectious diseases, where infections are usually under-reported, but inference of the true transmission dynamics is of utmost importance \cite{YanLipSha15}.

In control applications, individual dynamics are often estimated using for instance the Kalman filter \cite{Kal60}. In statistics, many methods are based on probabilistic graphical models (PGMs), e.g., the belief propagation algorithm \cite{Pea88}. However, neither approach is tractable when it comes to estimating the behavior of a large group of individuals simultaneously, let alone the fact that quite often only aggregate information is available. A number of methods have been developed to address this problem, for instance, the PGM framework has been extended to collective graphical models (CGMs), which is a formalism for inference and learning with aggregate data~\cite{SheDie11}. A CGM is a graphical model that describes the relationship between the aggregated counts of individuals. Several algorithms for aggregate marginal inference within the CGM framework have been proposed including Sinkhorn belief propagation (SBP)~\cite{SinHaaZha20}, approximate MAP inference~\cite{SheSunKumDie13} and non-linear belief propagation~\cite{SunSheKum15}. Whereas the latter two suffer from instability and lack convergence guarantees, the SBP algorithm is guaranteed to converge in case the underlying graphical model is acyclic. 

The SBP algorithm is based on multi-marginal optimal transport theory, which studies the problem of finding the most efficient transport plan between several distributions~\cite{Pas15,BenCarCut15}. It has been shown that the aggregate inference problem is equivalent to the entropic regularized formulation of a multi-marginal optimal transport problem \cite{SinHaaZha20}. With this equivalence, the celebrated Sinkhorn algorithm~\cite{Cut13,BenCarCut15} has been utilized to solve the aggregate inference problem.

Most estimation tasks in the control community involve dynamic systems and thus evolution over time. This dynamic nature adds a temporal component to CGM. As time evolves, the number of aggregate noisy observations and the size of the underlying graphical model increase constantly. Clearly, naive CGM inference algorithms such as SBP are not suitable for real-time operations as the computational complexity increases linearly with the length of the graph. 
Thus, to achieve real-time performance, we consider the problem of estimating the aggregate marginals of a Markov chain from noisy, aggregate observations in an incremental (online) fashion. In this problem, $M$ individuals behave independently according to the same underlying Markov chain for multiple time steps, and, at each time step, a noisy aggregate observation is made. As a new aggregate observation comes in at time step $t$, the goal is to estimate the aggregate marginal at time $t$. In case of an individual's model ($M=1$), this coincides with the traditional PGM, and naive incremental inference~\citet{Mur02} can be used by incorporating a sliding window filter to consider only a fixed number of most recent observations. 

For collective dynamics, following the naive incremental approach, we build on the SBP algorithm and propose the sliding window Sinkhorn belief propagation (SW-SBP) algorithm for the incremental (online) aggregate inference problem. The SW-SBP algorithm employs a sliding window filter of length $K$, i.e., at each time step the $K$ most recent observations are used in order to estimate the current hidden aggregate distribution. In order to capture previous information, we propose to add one node at the beginning of the window, for which we consider two different settings. In one setting, the marginal distribution of this node is specified. In the other one, the node introduces some prior potential. We evaluate the performance of our algorithm on a variety of scenarios, including population flow analysis, validating the working of our algorithm. 

The contribution of this work is twofold. On one hand, this is the first study of incremental inference in the CGM framework. Our method makes it possible to estimate the group behavior of a large collection of agents using aggregate observations in an online manner. On the other hand, this extends many incremental inference methods~\cite{Mur02,StaWhiBru17, WhiVu13} to the CGM setting. Indeed, when specialized to individual dynamics, our method reduces to a standard incremental inference algorithm.

\section{Background}\label{sec:background}
\subsection{Probabilistic Graphical Models and Belief Propagation} 
\label{subsec:pgm}
Consider a distribution of $J$ random variables with the same finite image space $\mathcal{X}$, where $|\mathcal{X}| = d$, and with some dependencies between them. A probabilistic graphical model (PGM) is a compact and intuitive representation of such a distribution, which describes the dependencies by a graph \cite{WaiJor08}. More precisely, for a graph $G=(V,\,E)$, the random variables are represented by the set of vertices $V$, and the dependencies by the set of edges $E$. 
The joint probability of the distribution of random variables can then be represented as
\begin{equation}\label{eq:individual}
p(\bx):= p(x_1,x_2,\ldots,x_J) = \frac{1}{Z} \prod_{(i,j) \in E} \psi_{ij}(x_i,x_j),
\end{equation}
where $\bx=\{x_1,\dots,x_J\}\in \mathcal{X}^J$, $\psi_{ij}\in\mathbb{R}$ are edge potentials, and $Z\in\mathbb{R}$ is a normalization constant. The edge potential $\psi_{ij}$ characterizes the strength of the dependency between the random variables at nodes $i$ and $j$. Sometimes one also defines node potentials $\phi_i(x_i)$, for $i\in V$. However, these can be absorbed in the edge potentials $\psi_{ij}$, and in this work we choose the compact notation \eqref{eq:individual}.

Often, one is interested in finding the distribution of one of the random variables, which is called the Bayesian marginal inference problem. An efficient method for this problem is the belief propagation algorithm, which updates the marginal distribution on the vertices, by sending messages (also called beliefs) between them \cite{Pea88}.
Let $N(i)$ denote the set of neighboring nodes of $i$. Then the message from variable node $i$ to variable node $j$ is
\begin{equation}\label{eq:BP}
m_{i \rightarrow j} (x_j) \propto \sum_{x_i} \psi_{ij}(x_i,x_j) 
\prod_{k\in N(i)\backslash j} m_{k \rightarrow i}(x_i).
\end{equation}
This message can be understood as the belief of node $i$ about node $j$.
The messages in \eqref{eq:BP} are updated iteratively over the graph. When the algorithm converges, the node and edge marginals are given by
\begin{subequations}\label{eq:belief}
	\begin{eqnarray}
	b_i(x_i) &\propto&  \prod_{k \in N(i)} m_{k \rightarrow i} (x_i)
	\\
	\!\! \!\! \!\! \!\! \!\! \!\! \!\! b_{ij} (x_i,x_j)\!\!\!\!\! &\propto&\!\!\!\!\! \psi_{ij}(x_i,x_j)\!\!\!\!\!\! \prod_{k\in N(i) \backslash j}\!\!\!\!\!\! m_{k\rightarrow i} (x_i) \!\!\!\!\!\!\! \prod_{\ell\in N(j) \backslash i} \!\!\!\!\!\! m_{\ell \rightarrow j} (x_j\!).
	\end{eqnarray}
\end{subequations}

For an acyclic graph the belief propagation algorithm converges globally \cite{YedFreWei01} and the estimated marginal distributions in \eqref{eq:belief} recover the true marginals exactly. Although convergence is not guaranteed for general graphs with cycles, in practice the belief propagation algorithm often performs well~\cite{MurWeiJor99}.

\subsection{Collective Graphical Models} \label{subsec:cgm}
Collective graphical models (CGMs) describe the distribution of the aggregate data from several populations, which are each sampled independently from a discrete graphical model \cite{SheDie11}.
Consider a PGM as in Section~\ref{subsec:pgm} with underlying graph $G=(V,E)$. 
Let $\bx^{(1)},...,\bx^{(M)}$ be $M$ samples of the PGM according to its joint probability distribution \eqref{eq:individual}. In particular, each sample is a tuple $\bx^{(m)}=(x_1^{(m)},\dots,x_J^{(m)})$, where $x_i^{(m)} \in \mathcal{X}$, for each $i=1,\dots,J$.

Let $X^{(m)}_i$ be the state of the $m^{th}$ individual at node $i$, and let $\mathbb{I}[.]$ denote the indicator function. Then the aggregate node distribution $\bn_i \in \mathbb{R}^d$ and aggregate edge distribution $\bn_{ij}\in \mathbb{R}^{d \times d}$ are element-wise given by
\begin{subequations}
\begin{align}
 n_i(x_i)=& \sum_{m=1}^M \mathbb{I}[X^{(m)}_i= x_i], \text{ for } i\in V,\\
    n_{ij}(x_i,x_j) =&  \sum_{m=1}^M \mathbb{I}[X_i^{(m)}= x_i, X_j^{(m)}= x_j], \text{ for } (i,j) \in E.
\end{align}
\end{subequations}
The collection of all the aggregate node and edge distributions is denoted as $\bn$, i.e., $\bn=\{\bn_i,\bn_{ij}| i\in V, (i,j)\in E\}$. 
By construction all entries of $\bn$ are integers and they satisfy
\begin{equation}\label{eq:nconstraints}
\begin{aligned}
    \sum_{x_i}n_i(x_i) &= M,\qquad\qquad\quad ~\text{ for } i \in V, \\
    n_i(x_i) &= \sum_{x_j} n_{ij} (x_i,x_j), \ \ \text{for } (i,j)\in E.
    \end{aligned}
\end{equation}

From the underlying PGM $p(\bx)$, one can calculate the probability distribution of $\bn$, and this is known as the CGM. Similar to PGM, an important problem in CGM is to infer the marginal distributions given some measurements.
Multiple algorithms for aggregate marginal inference within the CGM framework have been proposed including the recent Sinkhorn belief propagation (SBP)~\cite{SinHaaZha20}, approximate MAP inference~\cite{SheSunKumDie13} and non-linear belief propagation (NLBP)~\cite{SunSheKum15}.
The approximate MAP inference and NLBP often suffer from instability and lack of convergence, while SBP exhibits convergence guarantees for acyclic graphs.

\subsection{Sinkhorn Belief Propagation} \label{subsec:sinkhorn_bp}

The Sinkhorn belief propagation algorithm~\cite{SinHaaZha20} is based on belief propagation and utilizes the celebrated Sinkhorn algorithm for multi-marginal optimal transport~\cite{Nen16,Pas12,BenCarCut15}, in order to solve the aggregate inference problem efficiently.  

Let $G=(V,E)$ be a CGM, as in Section~\ref{subsec:cgm}, with joint aggregate distribution $\bn$. Moreover, let $\Gamma\subset\{1,\dots,J\}$ be a set of indices that represents all the nodes that are observed. That is, it holds $\bn_i= \by_i$ for a given set of observations $\by_i$, for $i\in\Gamma$ (see \cite{SinHaaZha20} for more discussion on this observation model).
In its variational form, the marginal inference problem for CGM with this observation model reads 
\begin{equation}\label{eq:agg_form}
\begin{aligned}
     \underset{\bn}{\text{min}} && {\rm KL}(\bn~ ||~ \prod_{(i,j)\in E} \psi_{ij}(x_i,x_j))\\
     \text{s. t.}  && \bn_i = \by_i, \quad \forall i \in \Gamma.
\end{aligned}
\end{equation}

When the underlying graph is a tree, the objective function of problem~\eqref{eq:agg_form} is the same as the Bethe free energy~\cite{YedFreWei05}
\begin{equation}\label{eq:Bethe_energy_MOT}
\begin{aligned}
    F_{\rm Bethe}(\bn) =  \sum_{i,j} \sum_{x_i,x_j} n_{ij}(x_i,x_j) \ln \frac{n_{ij}(x_i,x_j)}{\psi_{ij}(x_i,x_j)} \\
    - \sum_{i=1} (d_i - 1) \sum_{x_i} n_i(x_i) \ln n_i(x_i),
    \end{aligned}
    \end{equation}
together with the consistency constraints \eqref{eq:nconstraints}. Here, $d_i$ denotes the degree of node $i\in V$.
Then the aggregate inference problem, with aggregate observations $\by_i$ for $i \in \Gamma$, reads
\begin{subequations}\label{eq:MOT_bethe}
\begin{eqnarray}
    \min_{\bn_{ij},\bn_i} &&F_{\rm Bethe}(\bn) \label{eq:MOT_bethe_a}
    \\
   \text{s.t.} && n_i(x_i) = y_i(x_i),~ \forall i \in \Gamma \label{eq:MOT_bethe_b} \\
   && \sum_{x_j} n_{ij}(x_i,x_j) = n_i(x_i), \forall (i,j) \in E \label{eq:MOT_bethe_c}  \\
   && \sum_{x_i} n_i(x_i)= 1, ~ \forall i \in V.\label{eq:MOT_bethe_d}
\end{eqnarray}
\end{subequations}
Note that here \eqref{eq:MOT_bethe_b} corresponds to the aggregate observation constraints and \eqref{eq:MOT_bethe_c}-\eqref{eq:MOT_bethe_d} to the  normalized consistency constraints \eqref{eq:nconstraints}. The solution to this problem is characterized by the following result. 

\begin{thm}[{\cite[Theorem 1]{SinHaaZha20}} ] \label{thm:is_bp}
The solution to the aggregate inference problem~\eqref{eq:MOT_bethe} is characterized by 
\begin{equation}
    {n}_i (x_i)  \propto  \prod_{k \in N(i)}  m_{k \rightarrow i}(x_i), ~\forall i\notin \Gamma
\end{equation}
where $m_{i \rightarrow j}(x_j)$ are fixed points of
\begin{subequations}\label{eq:is_bp_MOT}
\begin{eqnarray}
    m_{i\rightarrow j} (x_j)  &\propto&  \sum_{x_i} \psi_{ij}(x_i,x_j) \prod_{k\in N(i)\backslash j}  m_{k\rightarrow i}(x_i); \nonumber \\
    &&\forall i \notin \Gamma,~  \forall j \in N(i), \label{eq:is_bp_MOT1}  \\
    m_{i\rightarrow j} (x_j) &\propto&  \sum_{x_i} \psi_{ij}(x_i,x_j) \frac{y_i(x_i)}{m_{j \rightarrow i} (x_i)}; \nonumber \\
    && \forall i \in \Gamma,~  \forall j \in N(i).  \label{eq:is_bp_MOT2}
\end{eqnarray}
\end{subequations}
\end{thm}
%

Note that the inference problem \eqref{eq:agg_form} can be viewed as an entropy regularized multi-marginal optimal transport problem \cite{BenCarCut15,Nen16,SinHaaZha20}.
As a result, one can utilize the efficient Sinkhorn algorithm to solve the inference problem \eqref{eq:agg_form}. This can be accelerated further by merging it with belief propagation. 
To summarize, by combining Sinkhorn for multi-marginal optimal transport problem and Theorem~\ref{thm:is_bp}, we establish Algorithm~\ref{alg:sbp}. We refer the reader to \cite{SinHaaZha20} for details of the derivation.
\begin{algorithm}[tb]
   \caption{Sinkhorn Belief Propagation (SBP)}
   \label{alg:sbp}
\begin{algorithmic}
   \STATE Initialize the messages $m_{i\rightarrow j} (x_j)$ 
   \STATE Update $m_{i\rightarrow j} (x_j)$ using \eqref{eq:is_bp_MOT}
   \WHILE{not converged}
   \FOR{$i \in \Gamma$}
        \STATE i) Update  $m_{i\rightarrow j} (x_j)$ using \eqref{eq:is_bp_MOT2}
        \STATE ii) Update all the messages on the path from $i$ to $i_{\rm next}$ 
        according to \eqref{eq:is_bp_MOT1}~(fix an order in $\Gamma$, $i_{\rm next}$ is the next element to $i$ in this order)
    \ENDFOR
    \ENDWHILE
\end{algorithmic}
\end{algorithm}
Note that in contrast to other algorithms for the aggregate inference problem, which rely on an explicit observation model~\cite{SunSheKum15}, SBP is guaranteed to converge when the underlying graph is a tree due to its foundations: Sinkhorn and belief propagation both converge \cite{YedFreWei01,KarRin17,BenCarCut15,ElvHaaJakKar20}.

The expressions in~\eqref{eq:is_bp_MOT} can be interpreted as messages between nodes, in analogy to the standard belief propagation method presented in Section~\ref{subsec:pgm}. In fact, the messages in \eqref{eq:is_bp_MOT} resemble the ones in \eqref{eq:BP}. The messages \eqref{eq:is_bp_MOT1} can be understood as a scaling step, which guarantees that the constraints \eqref{eq:MOT_bethe_b} remain satisfied.

\section{Main Results}\label{sec:main_result}
We consider \textit{incremental} marginal inference problems from aggregate data within the CGM framework. In particular, we address the problem of estimating the aggregate marginals of a hidden Markov chain from noisy aggregate observations in an incremental (online) fashion. We follow the model described in Section~\ref{subsec:sinkhorn_bp} as the generative model of aggregate data. A hidden Markov model (HMM) is a Markov chain where the state is not directly observable.
The joint distribution of an HMM factorizes as
\begin{equation} \label{eq:HMM_distribution}
    p(\bx, \bo) = p(x_1) \prod_{t=1} p(x_{t+1} \mid x_{t}) p(o_t \mid x_t),
\end{equation}
 where $\bx, \bo$ denotes state variable and observation respectively, $p(x_1)$ is the initial distribution of the starting state, $p(x_{t+1} \mid x_{t})$ are the transition probabilities between hidden (unobserved) variables, and $p(o_{t} \mid x_{t})$ are the observation probabilities for time steps $t=1,2,\ldots$. Note that the model given by \eqref{eq:HMM_distribution} is a directed graphical model but it can be equivalently converted to the undirected graphical model \eqref{eq:individual} by viewing the transition and observation probabilities as edge potentials $\psi_{ij}$ in \eqref{eq:individual}.

In this problem, $M$ individuals behave independently according to the same underlying hidden Markov chain for multiple time steps, and, at each time step, a noisy aggregate observation $\by_t$ is made in terms of the number of individuals in each observation node $o_t$. As a new observation is made at time step $t$, our goal is to estimate the corresponding aggregate marginal at time $t$. The number of aggregate noisy observations and the size of the underlying graphical model increases with time. To perform inference from this full data is not suitable for real-time operations as the computational complexity of inference increases greatly with the size of the graph. Therefore, incremental inference is of great importance in that it can efficiently estimate the required aggregate marginals from new observations without needing access to the previous observations.

\begin{figure}[t]
    \centering
    \includegraphics[scale = 0.5]{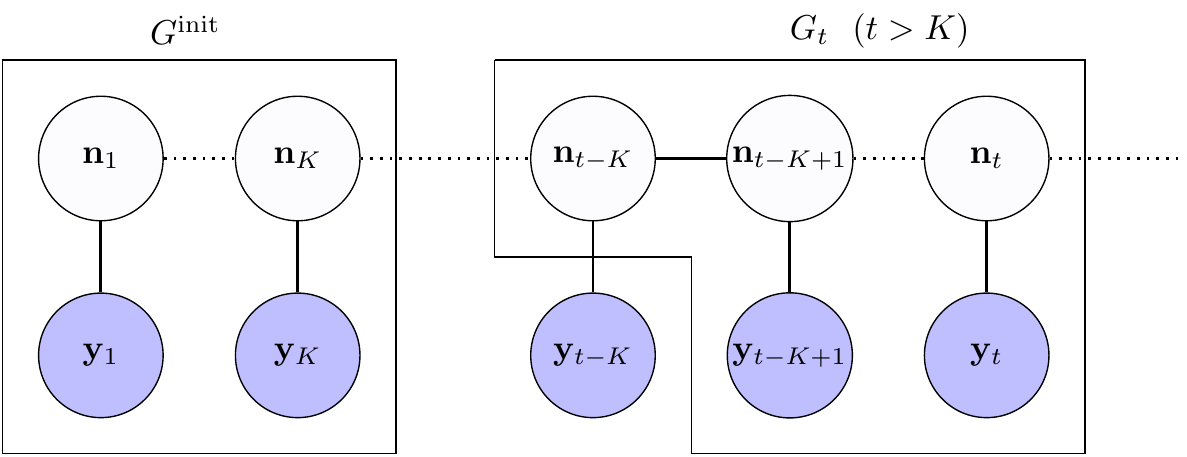}
    \caption{Incremental HMM in SW-SBP.}
    \label{fig:dynamic_markov}
\end{figure}

Building on the SBP algorithm, we propose a sliding window Sinkhorn belief propagation (SW-SBP) algorithm for the incremental (online) aggregate inference problem. Borrowing the idea from traditional incremental HMM inference \cite{Mur02}, the SW-SBP algorithm employs a sliding window filter of length $K$ so that at time step $t$ ($t>K$), only the $K$ most recent observations are used in order to estimate the conditional probabilities. In addition to using only $K$ previous observations, we propose to add a node%
\footnote{In case no extra node is added, the resulting algorithm is called naive SW-SBP.} to capture most of the information lost by discarding the previous observations. Denote this subgraph of the HMM model by $G_t = (V_t, E_t)$ as shown in Figure~\ref{fig:dynamic_markov}. Here, the incremental time starts at $t=K+1$. With this setting, we always have $2K + 1$ nodes (in $G_t$) at any time step $t$ from which the aggregate marginal is estimated. 
The Bethe free energy \eqref{eq:Bethe_energy_MOT} for the subgraph/sliding window $G_t$ with the underlying HMM model \eqref{eq:HMM_distribution} is
{\small
\begin{equation}\label{eq:Bethe_energy_HMM}
\begin{aligned}
     &\sum_{i=t-K+1}^{t} \bigg( \sum_{x_{i-1},x_{i}} n_{i-1,i}(x_{i-1},x_{i}) \ln \frac{n_{i-1,i}(x_{i-1},x_{i})}{p(x_{i}| x_{i-1})} + \\
     & \sum_{x_i, o_i} n_{i,i}(o_i,x_i) \ln  \frac{n_{i,i}(o_{i},x_{i})}{p(o_{i}| x_{i})}  - 2 \sum_{x_i} n_i(x_i) \ln n_i(x_i) \bigg) + \\
     & \sum_{x_t} n_t(x_t) \ln n_t(x_t) 
    \end{aligned}
\end{equation}}

We propose two methods for encoding information from the previous discarded observations in the node $\bn_{t-K}$. In the first method (SW-SBP-I) the node is constrained to be equal to the estimate of this node from the previous time step.
In the second method (SW-SBP-II) previous information is imposed by updating the potential $\phi_{t-K}$ on this node to the message $m_{t-K\to t-K+1}$ obtained at the previous time step.

\begin{figure}[t]
    \centering
    \begin{subfigure}[b]{0.19\textwidth}
        \centering
        \includegraphics[scale = 0.6]{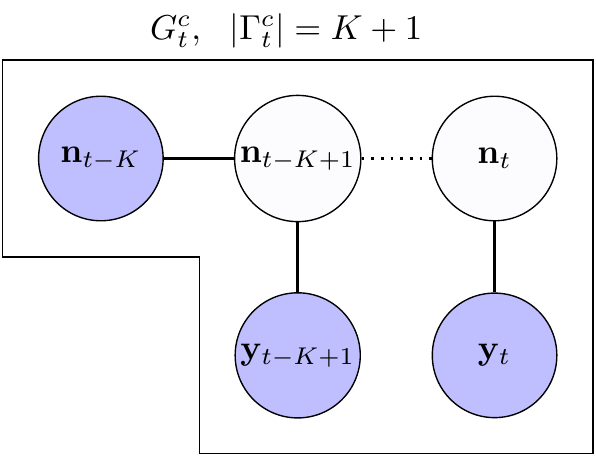}
        \caption{SW-SBP-I}
        \label{fig:Gt_constrained_potentialI}
    \end{subfigure} \hspace{0.1cm}
    \begin{subfigure}[b]{0.19\textwidth}
        \centering
        \includegraphics[scale = 0.6]{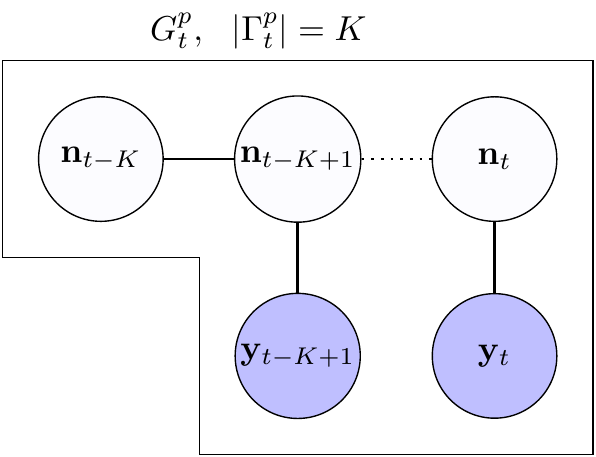}
        \caption{SW-SBP-II}
        \label{fig:Gt_constrained_potentialII}
    \end{subfigure}
    \caption{Incremental HMM graph for the two methods.}
     \label{fig:Gt_constrained_potential}
\end{figure}

 Figure~\ref{fig:Gt_constrained_potential} shows the incremental HMM graphs corresponding to the two proposed schemes. The graph corresponding to SW-SBP-I at time step $t$ is $G_t^c = (V_t^c,E_t^c)$ as shown in Figure~\ref{fig:Gt_constrained_potentialI}. Here, the node $\bn_{t-K}$ is a fixed marginal node, which equals the estimated marginal on $G_{t-1}^c$. The index set of constraints $\Gamma_t^c$ is described by the set of shaded nodes in Figure~\ref{fig:Gt_constrained_potentialI}. At time step $t$, SW-SBP-I solves the inference problem \eqref{eq:MOT_bethe} with $\Gamma= \Gamma_t^c$ and Bethe free energy given as in \eqref{eq:Bethe_energy_HMM}. The steps of SW-SBP-I are listed in Algorithm~\ref{alg:isbp_marginal}, where $G^{\text{init}}$ is the initial graph with the first $K$ number of observations ($2K$ nodes) as in Figure~\ref{fig:dynamic_markov}.

\begin{algorithm}[h]
   \caption{SW-SBP-I (constrained marginal)}
   \label{alg:isbp_marginal}
\begin{algorithmic}
   \STATE Run SBP on initial graph $G^{\text{init}}$
   \FOR{$t = K+1,K+2,\ldots,$}
        \STATE Constrain marginal $\bn_{t-K}$ to be equal to its estimate obtained from $G_{t-1}^c$
        \STATE Run SBP on $G_t^c$ 
    \ENDFOR
\end{algorithmic}
\end{algorithm}

For the SW-SBP-II case, the incremental graph at time $t$ is represented by $G_t^p = (V_t^p, E_t^p)$ as in Figure~\ref{fig:Gt_constrained_potentialII}.
Here, the index set of constraints $\Gamma_t^p$ does not contain $t-K$. Instead, the potential $\phi_{t-K}$ (which will be absorbed into $p(x_{t-K+1}\mid x_{t-K})$ and is thus not explicit in \eqref{eq:Bethe_energy_HMM}) is updated to be
the message from $\bn_{t-K}$ to $\bn_{t-K+1}$ on the graph $G_{t-1}^p$.
Algorithm~\ref{alg:isbp_potential} lists the steps of SW-SBP-II.

\begin{algorithm}[h]
   \caption{SW-SBP-II (potential update)}
   \label{alg:isbp_potential}
\begin{algorithmic}
   \STATE Run SBP on initial graph $G^{\text{init}}$
   \FOR{$t = K+1,K+2,\ldots,$}
        \STATE Update node potential $\phi_{t-K}$ by message $m_{t-K \rightarrow t-K+1}$ obtained from $G_{t-1}^p$
        \STATE Run SBP on $G_t^p$ 
    \ENDFOR
\end{algorithmic}
\end{algorithm}

\begin{rem} \label{rem:KF}
The idea to replace the previous observations with a single node potential as in SW-SBP-II has been widely adopted in standard filtering settings. For instance, a Kalman filter \cite{Kal60} utilizes this idea to efficiently estimate the current state for linear dynamic systems.
\end{rem}

\begin{rem}
When applied to a single HMM trajectory, that is, when all the measurements $\by_i$ are Dirac distributions, the SBP algorithm reduces \cite{SinHaaZha20} to the standard inference algorithm: Forward-backward algorithm \cite{WaiJor08} in HMM literature. This is due to the fact that when $\by_i$ is Dirac, the belief $m_{i\rightarrow j}$ it outputs is independent of what it receives by \eqref{eq:is_bp_MOT2}. Consequently, SW-SBP-II reduces to a standard incremental inference algorithm in HMM which also employs a potential node to capture discarded information when all the observations $\by_i$ are Diracs. 
\end{rem}

\section{Evaluation}\label{sec:eval}
We evaluate the performance of SW-SBP on multiple experiments including a toy example and a mobility pattern estimation problem.
We compare the performance of different SW-SBP algorithms in terms of the estimation error with the baseline algorithm, which is SBP (Algorithm \ref{alg:sbp}) on the whole graph without discarding previous measurements. 

\begin{figure}[t]
    \centering
    \begin{subfigure}[b]{0.47\textwidth}
        \centering
            \includegraphics[width=1.0\textwidth]{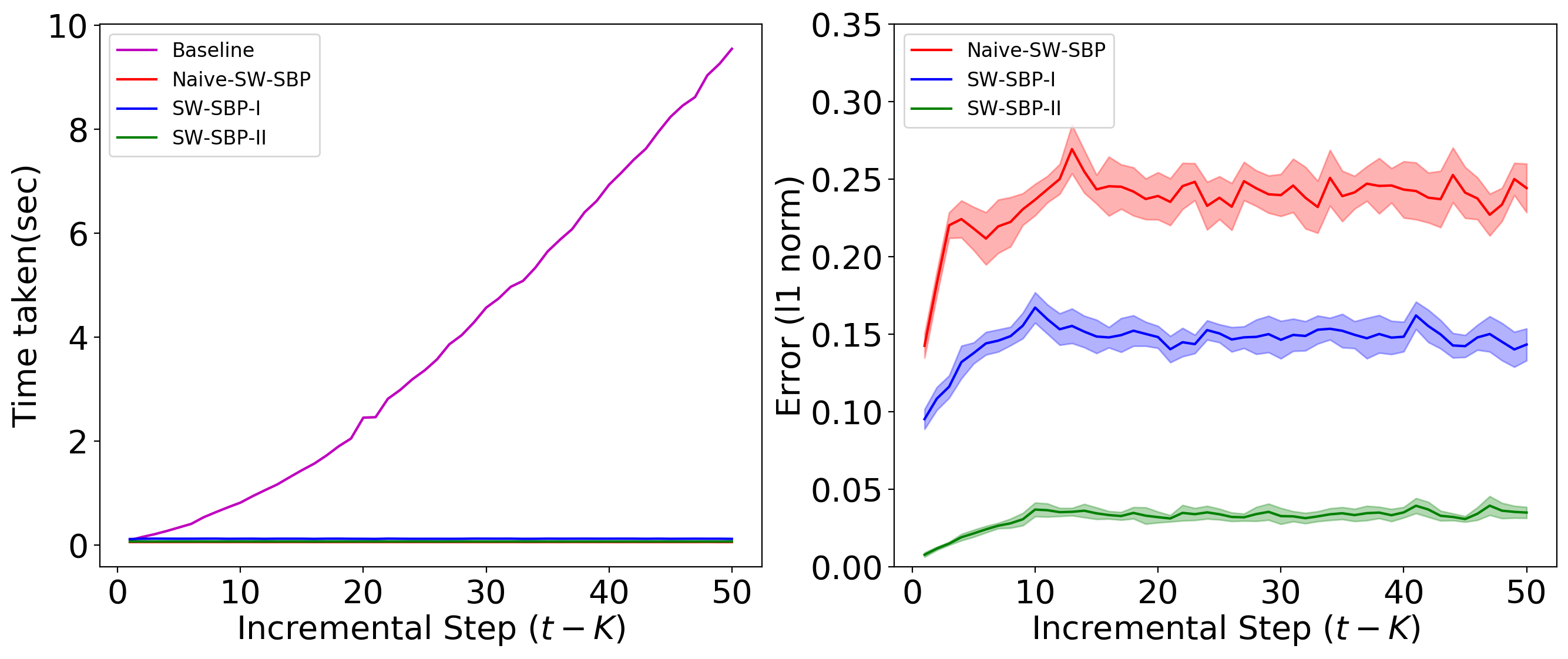}
            \caption{$K=3$.}
    \end{subfigure}
    \begin{subfigure}[b]{0.47\textwidth}
        \centering
            \includegraphics[width=1.0\textwidth]{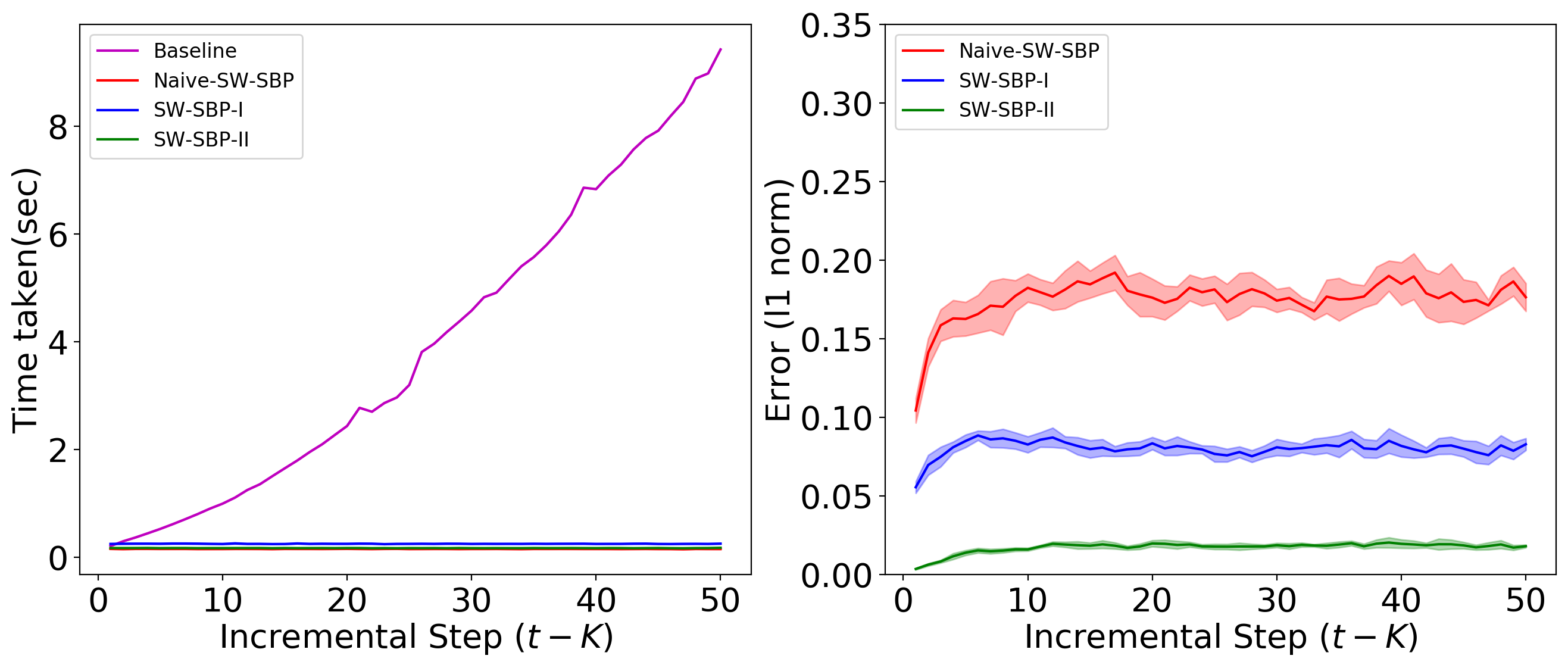}
            \caption{$K=5$.}
    \end{subfigure}
    \begin{subfigure}[b]{0.47\textwidth}
        \centering
            \includegraphics[width=1.0\textwidth]{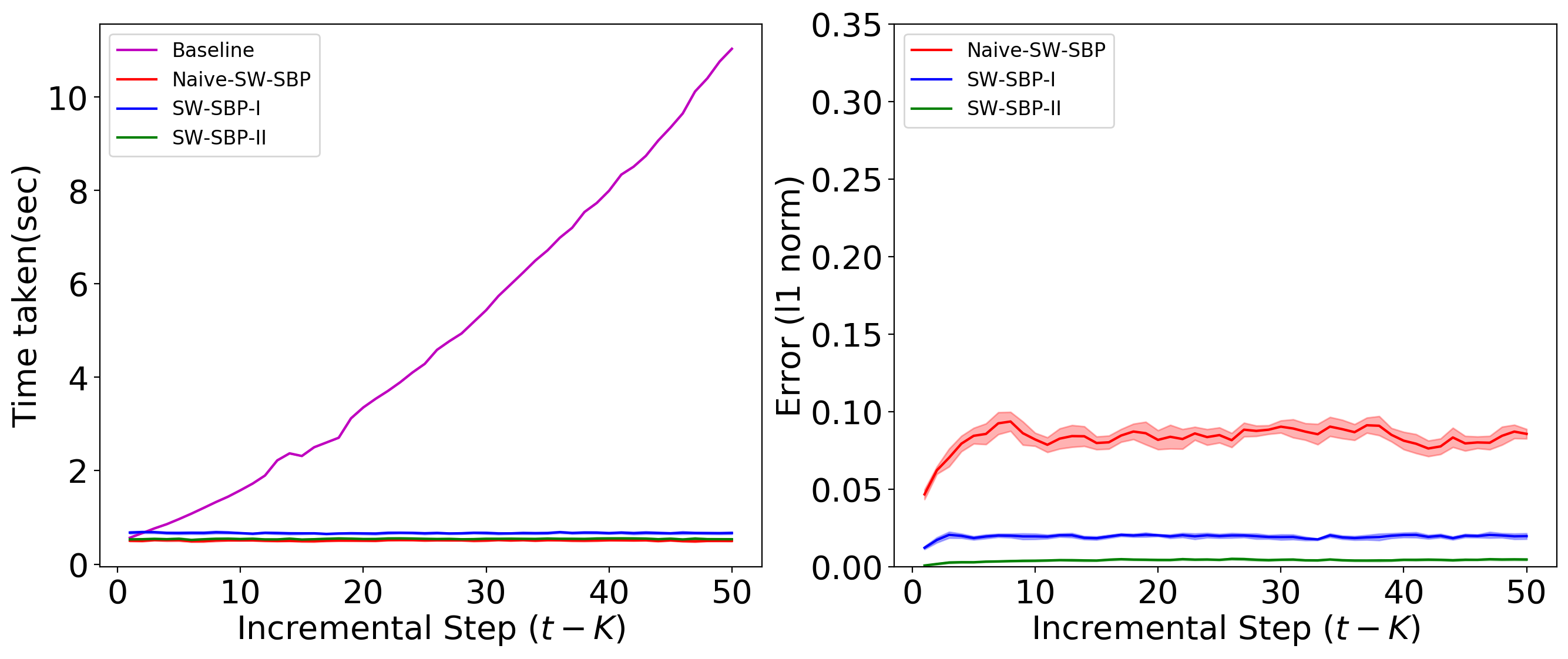}
            \caption{$K=10$.}
    \end{subfigure}
    \caption{Performance with $d =50$ for different window lengths $K$. The right column shows the $\ell_1$-norm error with respect to the baseline marginals and the left column shows the time consumption for the algorithms to converge. The solid lines denote the average over 10 trials and the shaded area represents the corresponding standard deviation.}
    \label{fig:comparison}
\end{figure}
\begin{figure}[t]
    \centering
    \begin{subfigure}[b]{0.21\textwidth}
        \centering
        \includegraphics[width=1.0\textwidth]{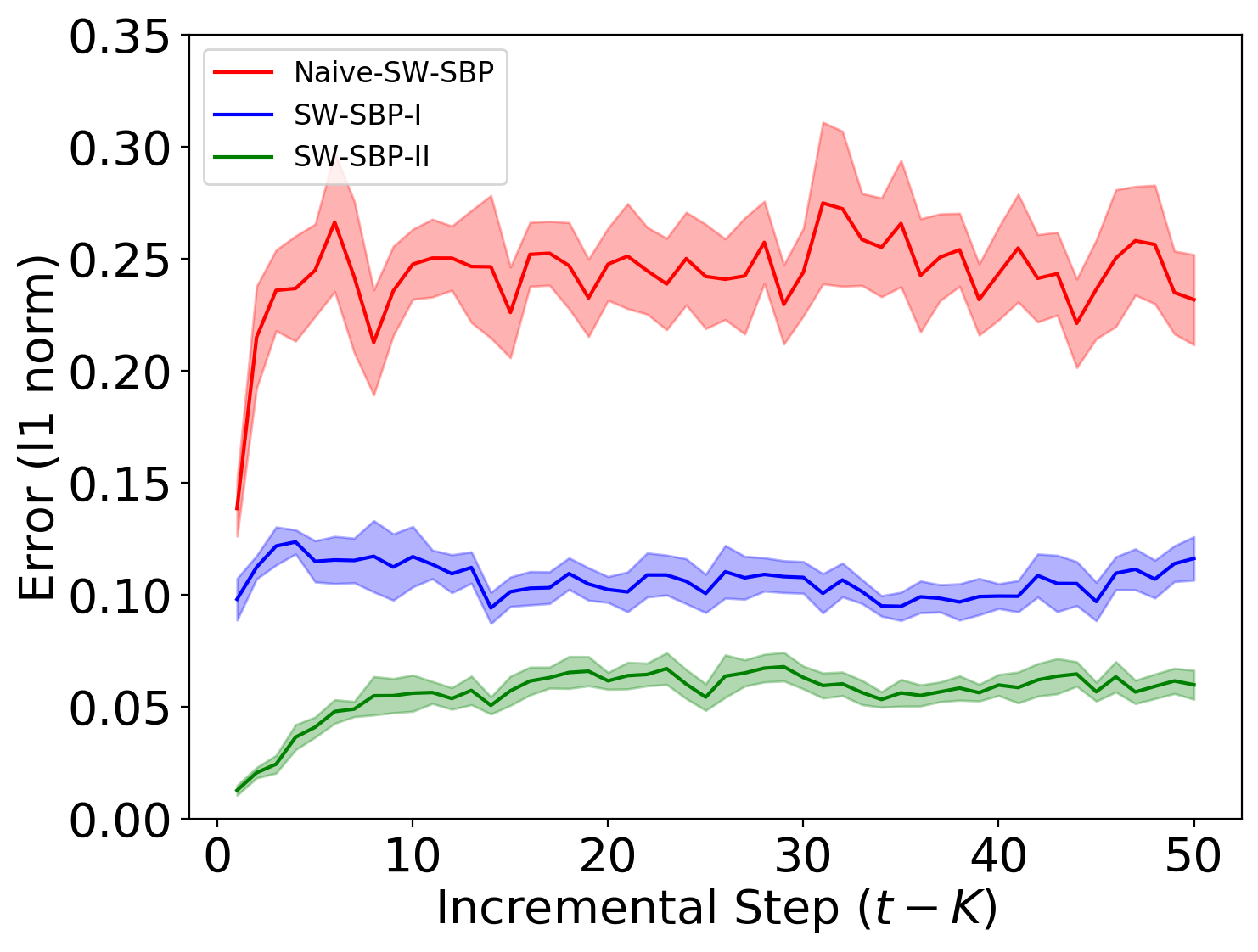}
        \caption{$d = 20$}
    \end{subfigure}
    \begin{subfigure}[b]{0.21\textwidth}
        \centering
        \includegraphics[width=1.0\textwidth]{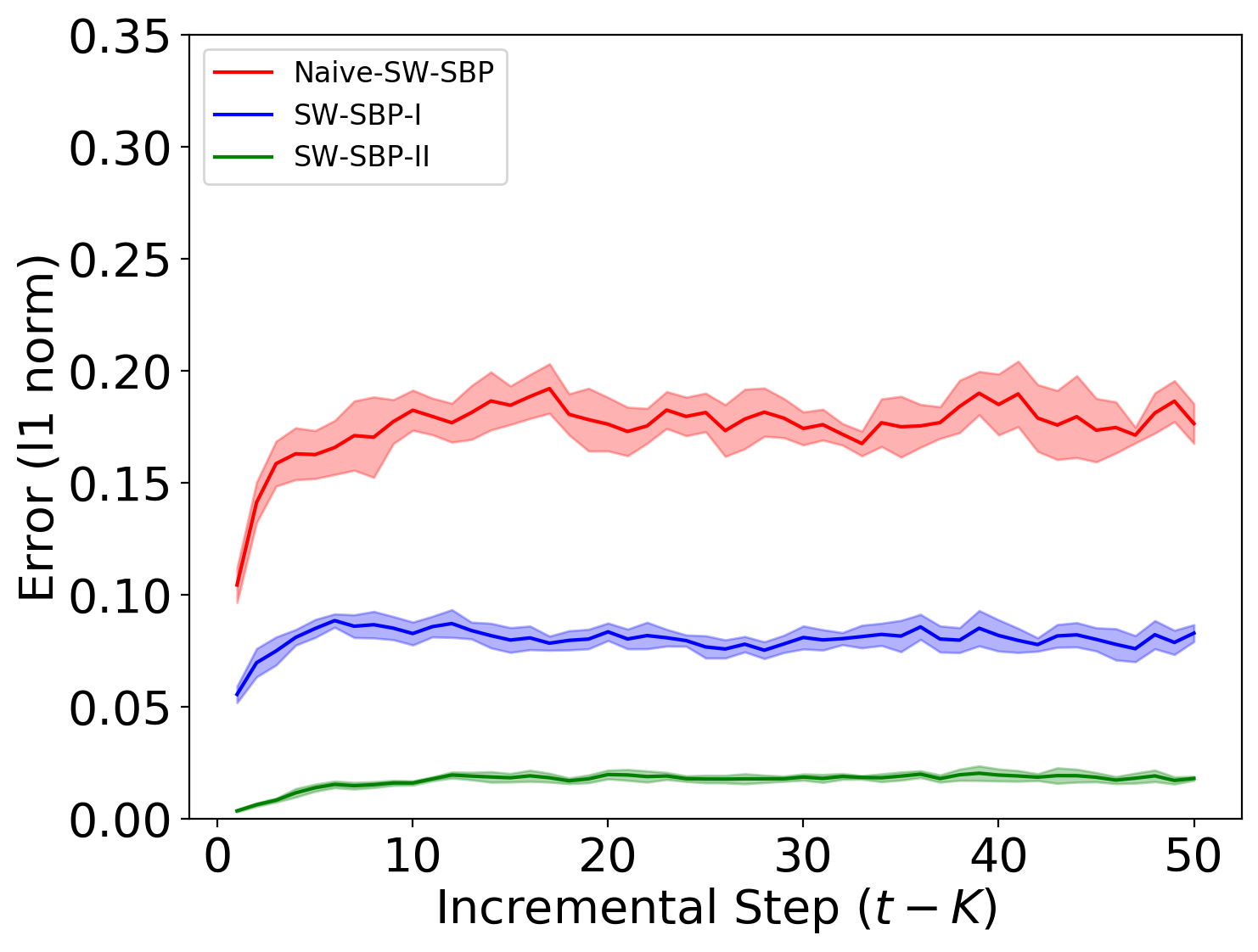}
        \caption{$d = 50$}
    \end{subfigure}
    \caption{Performance for different values of $d$ with fixed window length of $K=5$.}
    \label{fig:change_d}
\end{figure}

\subsection{Performance Comparison}
\label{subsec:toy}

To evaluate the performance of the proposed algorithms, we simulate a time-varying HMM with randomly chosen observations and evaluate the performance for different filter lengths $K$ and different values of state dimension $d$. The observation is of the same dimension as the state. The transition and observation matrices are both generated randomly using $500\mathcal{I} + 10\exp({E})$ where $\mathcal{I}$ denotes the identity matrix and $E$ is a random matrix generated from a standard Gaussian distribution.
Figure~\ref{fig:comparison} compares the performance of proposed methods for varying sliding window length. The results are averaged over 10 different trials. It is evident that the time consumption of the baseline method increases rapidly with the size of the graph. Moreover, the convergence behaviors of Naive-SW-SBP, SW-SBP-I, and SW-SBP-II are similar. In terms of estimation errors, SW-SBP-I and SW-SBP-II outperform Naive-SW-SBP by a significant margin. Moreover, SW-SBP-II shows better error performance as compared to SW-SBP-I and as the length of the filter increases, the errors decrease.
Figure~\ref{fig:change_d} depicts the error performance for different values of $d$ with fixed window length of $K=5$. It is observed from the figure that SW-SBP-I and SW-SBP-II perform similar for small values of $d$, outperforming Naive-SW-SBP algorithm, and SW-SBP-II performs better for large values of $d$. 
Note that the idea to update the potential in SW-SBP-II is inspired by the Kalman filter (see Remark~\ref{rem:KF}). In standard inference of HMMs, which is a special case of our framework where $\by_i$ is a Dirac distribution for all $i$, the strategy in SW-SBP-II gives precise solutions, that is, SW-SBP-II is a precise incremental implementation of SBP. However, in cases where the measurements $\by_i$ are general distributions, SW-SBP-II only approximates SBP. Currently there is no rigorous justification why SW-SBP-II outperforms SW-SBP-I in our general setting.

\begin{figure}[t]
    \centering
    \includegraphics[trim={0 9 0 8},clip,scale = 0.32]{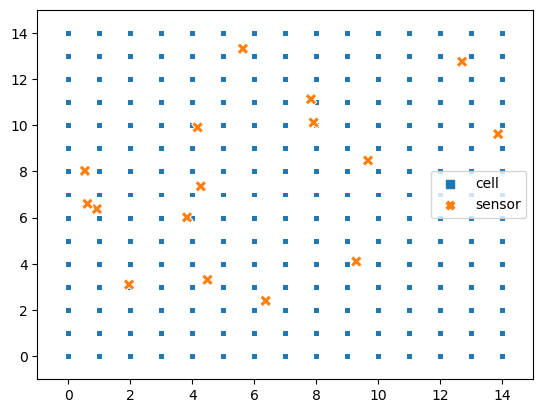}
    \caption{Location of $16$ sensors in a $15 \times 15$ grid.}
    \label{fig:sensor_locations}
\end{figure}

\begin{figure*}[t]
    \centering
     \begin{minipage}[c]{0.67\textwidth}
    \includegraphics[trim={50 0 60 5}, clip, width=\textwidth]{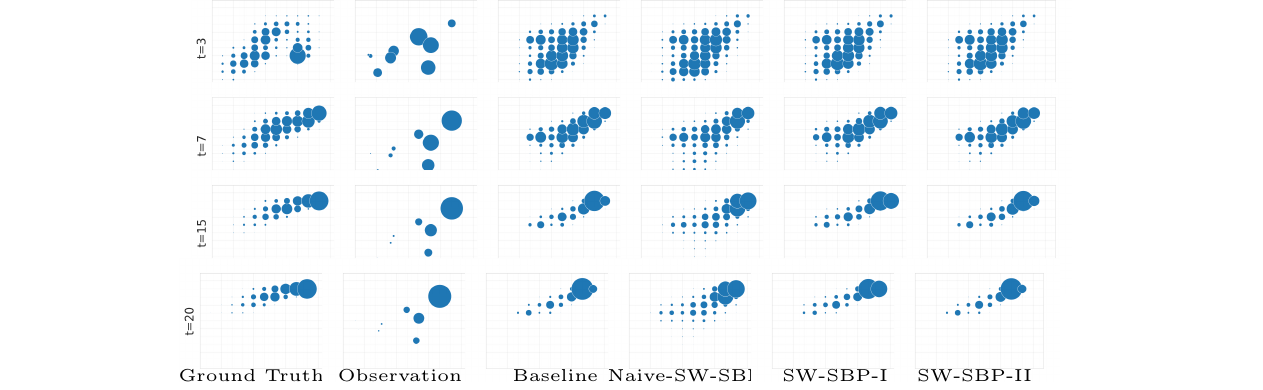}
     \end{minipage}
     \begin{minipage}[c]{0.32\textwidth}
    \caption{Simulation of movement of $M = 10000$ agents over $15 \times 15$ grid with $K = 3$. The first column depicts the real movement of agents at different time steps, second column represents the aggregate sensor observations, third column depicts estimated aggregated positions using the baseline full graph. Fourth, fifth, and sixth columns represent the estimated positions obtained by different variants of SW-SBP. The size of the circles is proportional to the number of agents.}
    \label{fig:human_mobility}
    \end{minipage} \vspace{-10pt}
\end{figure*}

\begin{figure}[t]
    \centering
    \begin{subfigure}[b]{0.42\textwidth}
        \centering
        \includegraphics[width=1.0\textwidth]{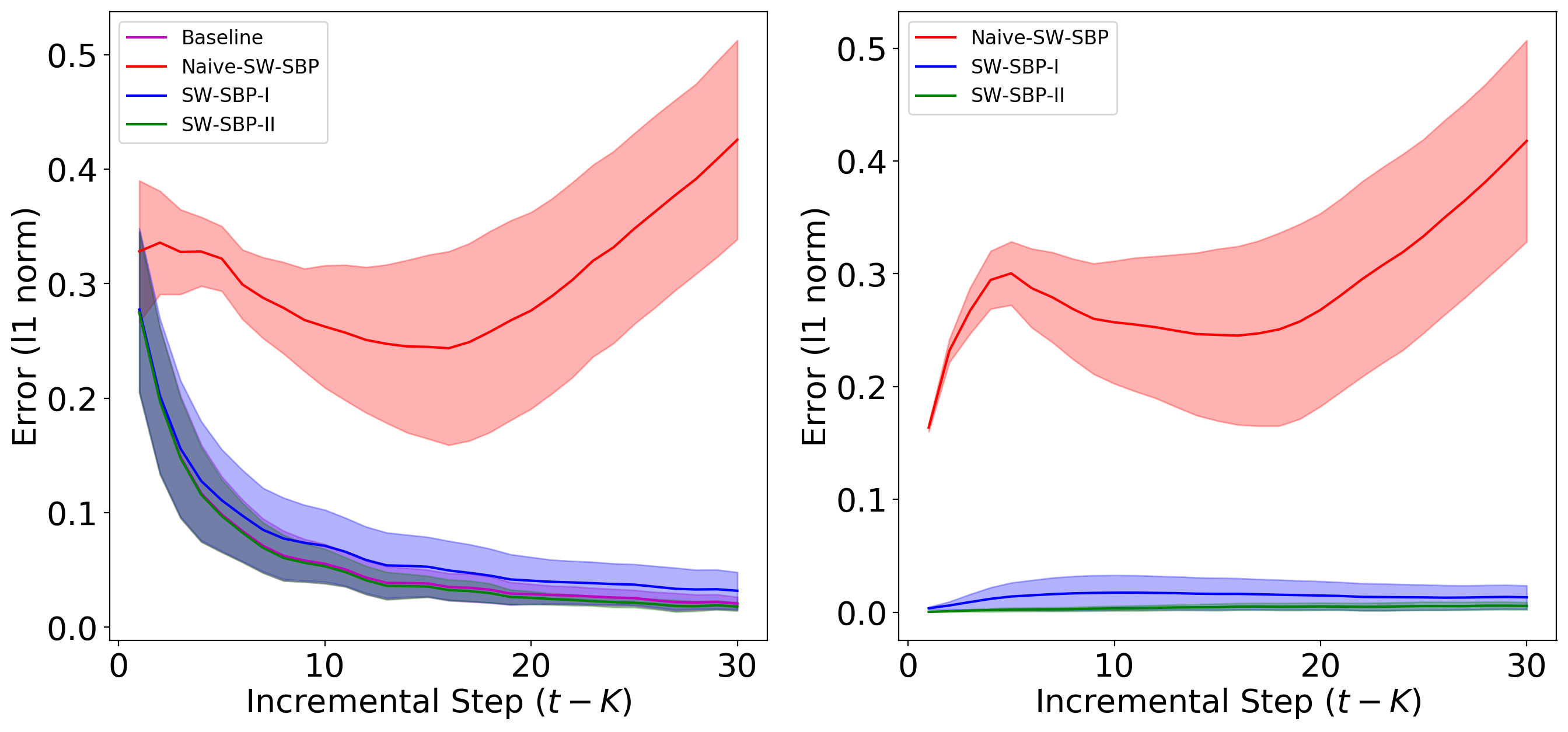}
    \end{subfigure}
    \caption{Error comparison for bird migration experiment. The left and right figure show the error with respect to the ground truth and with respect to the baseline estimate, respectively.}
    \label{fig:human_mobility_error}
\end{figure}

\subsection{Bird Migration} 
\label{subsec:human_mobility}
Next, we consider the problem of estimating a mobility pattern of birds in a geographical area from noisy aggregate counts. Following the environment considered in \citet{SunSheKum15}, we simulate $M$ individuals moving in an $L \times L$ grid, aiming to reach the top-right of the grid following a Markov chain. The transition probabilities of this Markov chain follow a log-linear distribution that accounts for four factors:  the distance between two positions, the angle between movement direction and the top-right direction, the angle between wind direction and the direction to goal (top-right in the grid), and preference to stay in the original cell. The weight for those four factors in the log-linear model is $(5,3,1.6,1)$. The observations are made as the counts of individuals connected to the randomly distributed sensors in the grid (see Figure~\ref{fig:sensor_locations}). These sensors constitute a distributed sensor network in which each sensor can only detect the present of birds nearby while is not able to output their locations.
We assume that each individual is connected to exactly one of the sensors and the probability of the connection decreases exponentially as the distance between agent and the sensor increases.
At time $t=0$, the population is concentrated at the two clusters: one at the left-bottom and one at the center-bottom.

We simulate the Markov chain for $M=10000$ individuals independently and perform incremental inference using SW-SBP algorithms. We compare the performance of the algorithms in terms of the $\ell_1$-norm of the difference between the estimated marginals and the baseline marginals. Figure~\ref{fig:human_mobility_error} shows the errors for the algorithm for $30$ time steps on a $15 \times 15$ grid with $16$ randomly distributed sensors. It can be observed from the figure that SW-SBP-II performs significantly better than SW-SBP-I and Naive-SW-SBP. Moreover, the errors for all the three methods decrease as the length of the sliding window filter increases. Figure~\ref{fig:human_mobility} shows the true and estimated movements of the the population using different methods. As can be seen, SW-SBP-II infers the population movements closest to the baseline.

\section{Conclusion}
\label{sec:conclusion}

In this paper, we proposed an algorithm for incremental inference from aggregate data in a Markov chain by employing a sliding window filter. The performance of proposed methods were demonstrated on multiple scenarios including estimation of bird mobility patterns. The present work focus on discrete setting and its extension to continuous state system is a future direction.

\bibliography{icml_mot}

\begin{thebibliography}{10}
\providecommand{\url}[1]{#1}
\csname url@rmstyle\endcsname
\providecommand{\newblock}{\relax}
\providecommand{\bibinfo}[2]{#2}
\providecommand\BIBentrySTDinterwordspacing{\spaceskip=0pt\relax}
\providecommand\BIBentryALTinterwordstretchfactor{4}
\providecommand\BIBentryALTinterwordspacing{\spaceskip=\fontdimen2\font plus
\BIBentryALTinterwordstretchfactor\fontdimen3\font minus
  \fontdimen4\font\relax}
\providecommand\BIBforeignlanguage[2]{{%
\expandafter\ifx\csname l@#1\endcsname\relax
\typeout{** WARNING: IEEEtran.bst: No hyphenation pattern has been}%
\typeout{** loaded for the language `#1'. Using the pattern for}%
\typeout{** the default language instead.}%
\else
\language=\csname l@#1\endcsname
\fi
#2}}

\bibitem{SheDie11}
D.~R. Sheldon and T.~G. Dietterich, ``Collective graphical models,'' in
  \emph{Advances in Neural Information Processing Systems}, 2011, pp.
  1161--1169.

\bibitem{SunSheKum15}
T.~Sun, D.~Sheldon, and A.~Kumar, ``Message passing for collective graphical
  models,'' in \emph{International Conference on Machine Learning}, 2015, pp.
  853--861.

\bibitem{Eve03}
G.~Evensen, ``The ensemble kalman filter: Theoretical formulation and practical
  implementation,'' \emph{Ocean dynamics}, vol.~53, no.~4, pp. 343--367, 2003.

\bibitem{LorNae11}
R.~J. Lorentzen and G.~N{\ae}vdal, ``An iterative ensemble {K}alman filter,''
  \emph{IEEE Transactions on Automatic Control}, vol.~56, no.~8, pp.
  1990--1995, 2011.

\bibitem{CheKar18}
Y.~Chen and J.~Karlsson, ``State tracking of linear ensembles via optimal mass
  transport,'' \emph{IEEE Control Systems Letters}, vol.~2, no.~2, pp.
  260--265, 2018.

\bibitem{HasRinChe19}
I.~Haasler, A.~Ringh, Y.~Chen, and J.~Karlsson, ``Estimating ensemble flows on
  a hidden {M}arkov chain,'' in \emph{58th IEEE Conference on Decision and
  Control}, 2019.

\bibitem{YanLipSha15}
W.~Yang, M.~Lipsitch, and J.~Shaman, ``Inference of seasonal and pandemic
  influenza transmission dynamics,'' \emph{Proceedings of the National Academy
  of Sciences}, vol. 112, no.~9, pp. 2723--2728, 2015.

\bibitem{Kal60}
R.~E. Kalman, ``A new approach to linear filtering and prediction problems,''
  1960.

\bibitem{Pea88}
J.~Pearl, ``Probabilistic reasoning in intelligent systems: Networks of
  plausible inference,'' \emph{Morgan Kaufmann Publishers Inc}, 1988.

\bibitem{SinHaaZha20}
R.~Singh, I.~Haasler, Q.~Zhang, J.~Karlsson, and Y.~Chen, ``Inference with
  aggregate data: An optimal transport approach,'' in \emph{Under Review},
  2020.

\bibitem{SheSunKumDie13}
D.~Sheldon, T.~Sun, A.~Kumar, and T.~Dietterich, ``Approximate inference in
  collective graphical models,'' in \emph{International Conference on Machine
  Learning}, 2013, pp. 1004--1012.

\bibitem{Pas15}
B.~Pass, ``Multi-marginal optimal transport: theory and applications,''
  \emph{ESAIM: Mathematical Modelling and Numerical Analysis}, vol.~49, no.~6,
  pp. 1771--1790, 2015.

\bibitem{BenCarCut15}
J.-D. Benamou, G.~Carlier, M.~Cuturi, L.~Nenna, and G.~Peyr{\'e}, ``Iterative
  {B}regman projections for regularized transportation problems,'' \emph{SIAM
  Journal on Scientific Computing}, vol.~37, no.~2, pp. A1111--A1138, 2015.

\bibitem{Cut13}
M.~Cuturi, ``Sinkhorn distances: Lightspeed computation of optimal transport,''
  in \emph{Advances in neural information processing systems}, 2013, pp.
  2292--2300.

\bibitem{Mur02}
K.~Murphy, ``Dynamic bayesian networks: Representation, inference and
  learning,'' \emph{PhD thesis, University of California}, 2002.

\bibitem{StaWhiBru17}
G.~Stamatescu, L.~B. White, and R.~Bruce-Doust, ``Track extraction with hidden
  reciprocal chains,'' \emph{IEEE Transactions on Automatic Control}, vol.~63,
  no.~4, pp. 1097--1104, 2017.

\bibitem{WhiVu13}
L.~B. White and H.~X. Vu, ``Maximum likelihood sequence estimation for hidden
  reciprocal processes,'' \emph{IEEE Transactions on Automatic Control},
  vol.~58, no.~10, pp. 2670--2674, 2013.

\bibitem{WaiJor08}
M.~J. Wainwright and M.~I. Jordan, ``Graphical models, exponential families,
  and variational inference,'' \emph{Foundations and Trends{\textregistered} in
  Machine Learning}, vol.~1, no. 1--2, pp. 1--305, 2008.

\bibitem{YedFreWei01}
J.~S. Yedidia, W.~T. Freeman, and Y.~Weiss, ``Generalized belief propagation,''
  in \emph{Advances in neural information processing systems}, 2001, pp.
  689--695.

\bibitem{MurWeiJor99}
K.~P. Murphy, Y.~Weiss, and M.~I. Jordan, ``Loopy belief propagation for
  approximate inference: An empirical study,'' in \emph{Proceedings of the
  Fifteenth conference on Uncertainty in artificial intelligence}.\hskip 1em
  plus 0.5em minus 0.4em\relax Morgan Kaufmann Publishers Inc., 1999, pp.
  467--475.

\bibitem{Nen16}
L.~Nenna, ``Numerical methods for multi-marginal optimal transportation,''
  Ph.D. dissertation, 2016.

\bibitem{Pas12}
B.~Pass, ``On the local structure of optimal measures in the multi-marginal
  optimal transportation problem,'' \emph{Calculus of Variations and Partial
  Differential Equations}, vol.~43, no. 3-4, pp. 529--536, 2012.

\bibitem{YedFreWei05}
J.~S. Yedidia, W.~T. Freeman, and Y.~Weiss, ``Constructing free-energy
  approximations and generalized belief propagation algorithms,'' \emph{IEEE
  Transactions on information theory}, vol.~51, no.~7, pp. 2282--2312, 2005.

\bibitem{KarRin17}
J.~Karlsson and A.~Ringh, ``Generalized {S}inkhorn iterations for regularizing
  inverse problems using optimal mass transport,'' \emph{SIAM Journal on
  Imaging Sciences}, vol.~10, no.~4, pp. 1935--1962, 2017.

\bibitem{ElvHaaJakKar20}
F.~Elvander, I.~Haasler, A.~Jakobsson, and J.~Karlsson, ``Multi-marginal
  optimal transport using partial information with applications in robust
  localization and sensor fusion,'' \emph{Signal Processing}, 2020.

\end{thebibliography}
\bibliographystyle{IEEEtran}

\end{document}